\documentclass{article}
\usepackage{graphicx}
\usepackage{subcaption}
\usepackage{graphicx}
\usepackage{enumitem}
\usepackage{PRIMEarxiv}
\usepackage{xcolor}
\usepackage[utf8]{inputenc} 
\usepackage[T1]{fontenc}    
\usepackage{hyperref}       
\usepackage{url}            
\usepackage{booktabs}       
\usepackage{amsfonts}       
\usepackage{nicefrac}       
\usepackage{microtype}      
\usepackage{lipsum}
\usepackage{fancyhdr}       
\usepackage{graphicx}       
\graphicspath{{media/}}     
\usepackage{amsmath}
\usepackage{algorithm}
\usepackage[noend]{algpseudocode}

\title{Adversary ML Resilience in Autonomous Driving through Human Centered Perception Mechanisms
}

\author{
  Aakriti Shah \\
  Rollins College \\
  ashah@rollins.edu
}

\begin{document}
\maketitle

\vspace{-0.75cm} 

\begin{abstract}
Physical adversarial attacks on road signs are continuously exploiting vulnerabilities in modern day autonomous vehicles (AVs) and impeding their ability to correctly classify what type of road sign they encounter. Current models cannot generalize input data well, resulting in overfitting or underfitting. In overfitting, the model memorizes the input data but cannot generalize to new scenarios. In underfitting, the model does not learn enough of the input data to accurately classify these road signs. This paper explores the resilience of autonomous driving systems against three main physical adversarial attacks (tape, graffiti, illumination), specifically targeting object classifiers. Several machine learning models were developed and evaluated on two distinct datasets: road signs (stop signs, speed limit signs, traffic lights, and pedestrian crosswalk signs) and geometric shapes (octagons, circles, squares, and triangles). The study compared algorithm performance under different conditions, including clean and adversarial training and testing on these datasets. To build robustness against attacks, defense techniques like adversarial training and transfer learning were implemented. Results demonstrated transfer learning models played a crucial role in performance by allowing knowledge gained from shape training to improve generalizability of road sign classification, despite the datasets being completely different. The paper suggests future research directions, including human-in-the-loop validation, security analysis, real-world testing, and explainable AI for transparency. This study aims to contribute to improving security and robustness of object classifiers in autonomous vehicles and mitigating adversarial example impacts on driving systems.
\end{abstract}


\footnotetext[1]{This research was conducted at Pennsylvania State University through a National Science Foundation funded \lq Research Experiences for Undergraduates' (REU) program. The research and paper writing process took ten weeks, from late May to early August of 2023.}

\section{Introduction}
\hspace{1cm} The rapid development of autonomous driving technology has paved the way for safer and more efficient transportation systems. However, with the integration of Machine Learning (ML) algorithms in SAE Level 3 autonomous vehicles (AVs), new challenges have emerged in the form of adversarial attacks, in which the systems have become less safe, robust, and effective. These attacks exploit vulnerabilities in ML models and lead to potential risks to the safety and accuracy of autonomous driving systems. Specifically, this research dives into the investigation of the possibility of integrating human perception properties in order to mitigate the effects of these physical adversarial examples.
\subsection{Problem}
\hspace{1cm} The main purpose of this research is to address the vulnerabilities caused by physical adversarial attacks in modern day AVs. These attacks are aimed at exploiting vulnerabilities in models and involve perturbing input data to mislead the model’s predictions. These attacks can interfere with the system’s awareness of its environment, something which is intuitively crucial for self-driving vehicles. Therefore, developing dependable and secure defense techniques to enhance the robustness of the system against real-time adversarial attacks is essential. The solutions and current literature surrounding this problem are unable to generalize the input data they are given, which allows them to perform well on the data they were trained on, but perform poorly on unseen (testing) data. This often results in overfitting, limited applicability and poor performance in the real world. This research aims to strengthen object classifier systems, particularly by employing Convolutional Neural Networks (CNNs) and human perception mechanisms, and enhancing human-vehicle interfaces to effectively mitigate the impact of adversarial attacks.
\subsection{Objectives}
This research is guided by the following objectives:
\begin{itemize}
\item Analyze the impact of physical adversarial attacks on autonomous driving systems and assess the potential risks they pose to passengers, pedestrians, and other vehicles.
\item Explore and develop defense techniques, leveraging CNNs, to detect and mitigate physical adversarial examples in real-time autonomous driving situations.
\item Investigate the application of transfer learning algorithms, particularly CNN-based models, to improve the adaptability and generalizability of autonomous driving models, considering the problems caused by physical adversarial attacks.
\item Enhance human-vehicle interfaces to provide drivers with insights into the state of the system and leverage their input to reinforce the system's robustness in the presence of adversarial attacks.
\end{itemize}
\subsection{Methodology Overview}
\hspace{1cm} To achieve the research objectives, an experimental approach will be used. Different algorithms, which incorporate defense techniques such as adversarial training, as well as the concept of transfer learning and human perception properties, will be evaluated and compared in an effort to make an object detector that is able to generalize and be more robust against adversarial attacks. There will be 6 different algorithms, all using a multi-class classifier and testing the effects that different datasets and their exposures to 3 main adversarial examples – tape, graffiti, and illumination – have on the model’s ability to be able to accurately detect which road sign is present (stop sign, speed limit sign, pedestrian crosswalk sign, traffic light), irrespective of whether there is an attack present. This research will focus on testing against various small-scale scenarios with minimal attack types, which should not be extrapolated to life-or-death situations. The algorithms will be compared based on their accuracy, ability to generalize, and computational efficiency.
\subsection{Motivation and Limitations}
\hspace{1cm} The motivation behind this research lies in addressing the vulnerabilities of current machine-learning algorithms and their inability to detect road signs, when there is some sort of physical adversarial attack present. These attacks can pose a serious threat to public safety and simultaneously undermine the public’s trust in autonomous vehicle technology. By understanding the vulnerabilities and developing a more robust defense algorithm, this research contributes to the development of safe and reliable AV systems. This research aims to empower drivers to provide input when presented with adversarial attacks, creating a more robust system by utilizing human input and oversight when needed. The findings from this study will offer insights into potential strategies for future research. These include exploration into more diverse test scenarios, the utilization of more extensive datasets to enhance model generalization and effectiveness, and the incorporation of additional security measures to address human input vulnerabilities more effectively.
\par 
\hspace{1cm} Special considerations will be made to protect data privacy. Collecting sufficient data for training autonomous driving models requires acquiring vast amounts of real-world data, which can be a significant limitation. Ensuring this training data is not skewed, obscured, or augmented in an unsuitable or inauspicious manner for the purposes of investigation is of the utmost importance to protect its integrity and reliability. Since autonomous driving systems could potentially encounter life-or-death situations, it is important to note that this investigation has been carefully examined within the context of small-scale scenarios with limited test data, serving as an initial exploration into the broader scheme of autonomous vehicle research.
\par 
\hspace{1cm} The scope of this research entails an investigation of physical adversarial attacks on autonomous driving systems, employing CNNs as the primary deep learning algorithm to strengthen object classifiers. This will be done by evaluating and comparing current algorithms to mitigate adversarial attacks, utilizing tactics that humans use to learn road signs and employing them, and finally comparing these algorithms and evaluate them against current techniques. However, due to the evolving nature of adversarial attacks and time restraints, it is not be possible to address all potential attack scenarios comprehensively, and can only be tested under a hand-full of scenarios. There is a limitation of viable image, sensor, and traffic data that is publicly available to train these models. Additionally, the research will focus on strengthening algorithms, models, and human-vehicle interfaces, but it will not delve into the broader security risk that is implied by human-computer interaction, i.e. human-in-the-loop-validation. This human involvement allows for human input to be directly added into the training data of the model, which can possibly make the model biased, or just incorrect in general. Due to time restraints, the security of this added feature will be developed at a later time.
\subsection{Research Question}
How can defense techniques and transfer learning algorithms enhance the robustness of autonomous driving systems against physical adversarial attacks in real time and strengthen human-vehicle interfaces?

\section{Related Works}
\subsection{Adversarial Examples in Deep Learning}
\hspace{1cm} According to Ian Goodfellow et al. \cite{goodfellow}, adversarial examples are input samples that are carefully crafted to deceive machine learning models. They state, “adversarial examples are inputs to machine learning models that an attacker has intentionally designed to cause the model to make a mistake” \cite{goodfellow}. The papers in this section explore the generation, impact, and defense against adversarial examples. Shang \cite{shapeshifter} introduces ShapeShifter, a physical adversarial attack method that can fool object detectors in the real world, while Dawn investigates physical adversarial examples that deceive object detectors in real-world scenarios. They mention they “demonstrate that physical adversarial examples can be created to deceive object detectors in real-world scenarios” \cite{shapeshifter}. This presents the need for modern systems in autonomous vehicles to be aware of adversarial attacks and equipped with the resources they need to handle run ins with them in the correct manner. Adachi \cite{song} proposes a novel adversarial training method that combines masking and mixing techniques to improve the robustness of deep learning models against adversarial attacks.

\subsection{Attacks and Situational Awareness in Computer Vision}
\hspace{1cm} The papers related to Attacks and Situational Awareness in Computer Vision focus on remote perception attacks in computer vision systems. Yanmao Man \cite{man} explores the vulnerabilities of camera-based image classification systems to remote perception attacks. They discuss different attack scenarios and techniques and discuss their investigation into “remote perception attacks against camera-based image classification systems, including adversarial perturbations and physical-world attacks” \cite{man}. Katherine Zhang's \cite{aiping} paper examines the situation awareness of human drivers in the presence of physical-world attacks on autonomous driving systems. They mention that they “study the situation awareness of human drivers under physical-world attacks on autonomous driving systems” to investigate the effects of different attack scenarios on human perception and decision-making \cite{aiping}.

\subsection{Color and Shape Psychology}
\hspace{1cm} These papers focus on the application of color psychology in visual perception and design. Simona Buetti \cite{buetti} explores how color and shape combine in the human visual system to direct attention. They investigate the relationship between color, shape, and attention, discussing how “color and shape combine in the human visual system to direct attention” \cite{buetti}. Jicheng Yang and Xiaoying Shen \cite{yang} examine the application of color psychology in community health environment design. They discuss how color choices can impact mood, well-being, and behavior in healthcare settings. This leads to the consideration of how color and shape can be used to attract attention to road signs, as humans are naturally drawn to specific colors and shapes.

\section{Methodology}
\subsection{Theoretical Framework}
\hspace{1cm} The theoretical foundation of This research is grounded in concepts of defense techniques against adversarial attacks, transfer learning, and the psychology behind how humans learn. Adversarial examples are "inputs formed by applying small but intentionally
worst-case perturbations to examples from the dataset, such that the perturbed input results in the model outputting an incorrect answer with high confidence" \cite{goodfellow}. They are employed to intentionally deceive machine learning models. With the prevalence of these algorithms in modern day technologies, defense techniques are put into place in order to enhance the robustness of these models against such attacks. One of which being adversarial training, a common defense mechanism that is used to train models on both clean and adversarial examples to enhance their strength and mitigate the damage that can be done by the attacks. It can also "provide an additional regularization benefit beyond that provided by using generic regularization strategies" like normalization, pretraining, and adding dropout layers \cite{goodfellow}. Another of these defense techniques is a method known as transfer learning, which leverages knowledge that is learned from one task in order to improve its performance on another task, improving its ability to generalize \cite{TL}.

\subsection{Human Perception Mechanisms}
\hspace{1cm} Human perception is a complex process that enables humans to navigate and interpret their environment. At the core of this cognitive ability lies the visual perception system, where the brain processes the incoming sensory information from the eyes to identify and recognize objects. An important aspect of this process involves the utilization of shapes and colors as primary cues for object detection and classification. The brain's ability to determine objects based on their unique shapes allows humans to recognize familiar objects even when they are partially occluded or viewed from different angles \cite{gov}.

\par \hspace{1cm} Road sign learning demonstrates the importance of shapes and colors in practical scenarios. From an early age, individuals are exposed to road signs that use simple and recognizable shapes and color patterns. This early exposure and the brain's innate ability to associate specific shapes and colors with particular meanings and emotions lead to efficient learning and quick comprehension of road signs \cite{yang}. For instance, the octagonal shape of a stop sign and its bright red color instantly convey the message to drivers that they need to halt their vehicles. Similarly, the triangular shape and bold yellow color of warning signs alert drivers to potential hazards ahead.

\par \hspace{1cm} Inspired by these human perception mechanisms, I propose a strategy that aims to harness the strengths of shapes and colors in object detection and road sign classification. By integrating shape-based representations and leveraging color contrasts, the algorithm seeks to achieve robust and efficient object recognition, particularly in the context of road signs. I anticipate that this approach will lead to improved performance and better generalization in real-world scenarios. I acknowledge that it may have its limitations, especially when attempting to replicate the intricacies of human perception entirely. Nevertheless, I believe that our research may open up more avenues for further exploration to contribute to the advancement of computer vision technologies that closely resemble human-like perception and our understanding of the world.

\subsection{Algorithm Design and Implementation}
\hspace{1cm} Due to the necessity of conducting a comparative analysis, six algorithms were implemented for the purpose of road sign recognition in this experimental study. Each algorithm utilizes a deep learning model with a Convolutional Neural Network (CNN) architecture. The algorithms differ in their training and testing data, allowing us to evaluate their performance under various scenarios. 

\subsubsection{Algorithm 1: Control Algorithm}
\hspace{1cm} The control algorithm serves as the baseline for performance evaluation. It is trained and tested on clean road sign data and aims to classify these road signs into their respective categories: stop sign, speed limit sign, traffic light, and crosswalk sign. The CNN model consists of one input layer, three convolutional layers, three max pooling layers, one flatten layer, two dense layers, and one output layer. This neural network structure will remain consistent for all 6 algorithms. The design of this model allows it for the learning of relevant features from the input images, so the model is able to make accurate predictions in real-time. The clean bounding box labeled dataset is split into 3 sections, with 70\% of it being training data, 20\% validation, and 10\% testing. Under normal driving circumstances, Algorithm 1 is expected to achieve the highest accuracy among the algorithms.

\subsubsection{Algorithm 2: Adversarial Road Signs Algorithm}
\hspace{1cm} The second algorithm focuses on the robustness of the model by training it on clean road sign data and testing it on adversarial road sign data. The CNN of this model has the same architecture as that of the first algorithm. The purpose of this algorithm is to evaluate the performance of the model when faced with adversarial road sign inputs that may contain perturbations or modifications designed to deceive the model, with no prior exposure to these adversarial examples. Just as Algorithm 1, the bounding box labeled dataset is split into 3 sections, but is split for a second time in order to account for the adversarial examples that must be incorporated into the testing and validation sets. In this algorithm, the training set, validation set, and testing set are still the 70\%, 20\%, and 10\%, respectively, as they were in Algorithm 1. The validation and testing sets are split again so that one half of their clean data being set aside to be augmented. This leaves the clean training data alone, and just augments 50\% of the validation and testing data so that adversarial examples, along with clean examples are incorporated. Under both normal and adversarial driving circumstances, Algorithm 2 is expected to achieve the lowest accuracy among the algorithms. Here are some examples of road signs with the 3 adversarial augmentations (tape, graffiti, and illumination) overlaid on them:

\begin{figure} [H]
  \centering
  \begin{minipage}{0.25\textwidth}
    \centering
    \includegraphics[width=\textwidth]{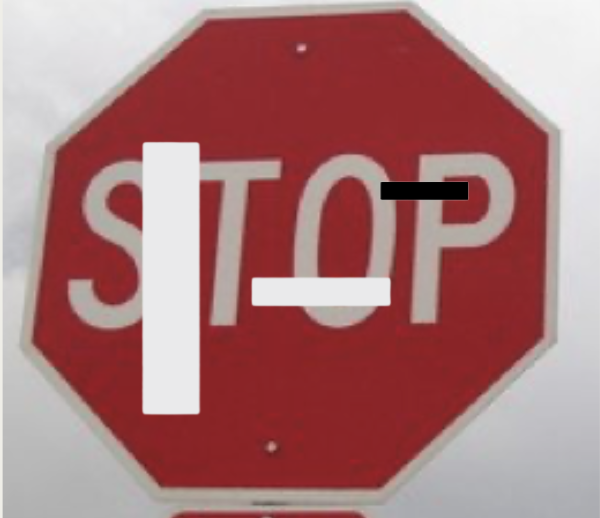}
    \caption{Tape Attack}
    \label{fig1:image1}
  \end{minipage}\hfill
  \begin{minipage}{0.25\textwidth}
    \centering
    \includegraphics[width=\textwidth]{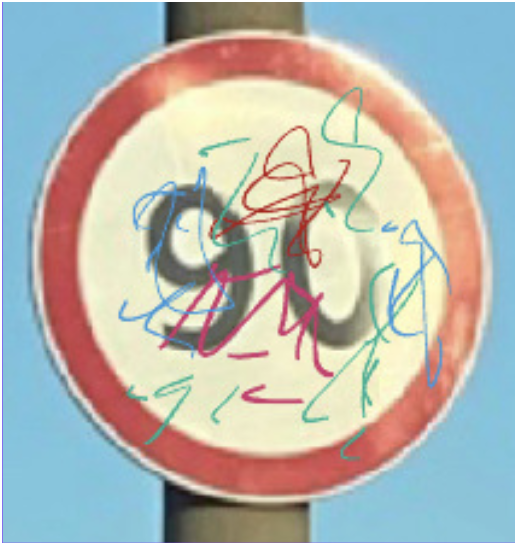}
    \caption{Graffiti Attack}
    \label{fig2:image2}
  \end{minipage}\hfill
  \begin{minipage}{0.25\textwidth}
    \centering
    \includegraphics[width=\textwidth]{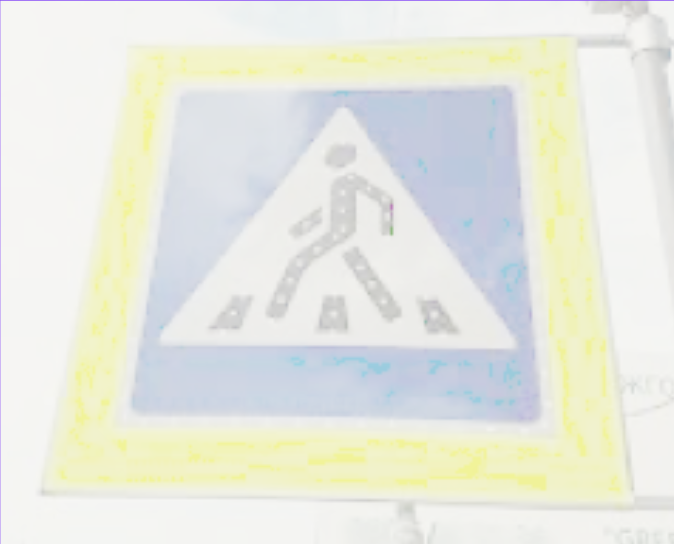}
    \caption{Illumination Attack}
    \label{fig3:image3}
  \end{minipage}
\end{figure} 

\subsubsection{Algorithm 3: Adversarial Road Signs with Adversarial Training Algorithm}
\hspace{1cm} The third algorithm builds on Algorithm 2 by incorporating adversarial training to improve the model's robustness against adversarial examples. The model is trained, validated, and tested on adversarial road sign data. This algorithm further investigates the effectiveness of adversarial training in enhancing the model's resistance to adversarial attacks as they apply to road signs. The CNN architecture remains the same as in the previous algorithms. Algorithm 3 is expected to outperform the other algorithms when facing adversarial attacks. The training set, validation set, and testing set are still the 70\%, 20\%, and 10\%, respectively, as they were in Algorithm 1 and 2. The training, validation, and testing sets are split again so that one half of each of their allotted clean is set aside to be augmented. 

\subsubsection{Algorithm 4: Clean Shapes with Clean Road Sign Algorithm}
\hspace{1cm} The fourth algorithm investigates the model's performance when trained on clean shapes data and tested on clean road sign data. This algorithm aims to evaluate the model's ability to recognize road signs based on their shapes alone, without the presence of adversarial perturbations. The CNN architecture remains the same as in the previous algorithms. The training dataset consists of clean shapes data, while the testing dataset comprises clean road sign images. Here are some examples of clean shape data:

\begin{figure} [H]
  \centering
  \begin{minipage}{0.25\textwidth}
    \centering
    \includegraphics[width=\textwidth]{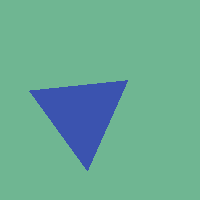}
    \caption{Triangle}{(Pedestrian Crosswalk Sign)}
    \label{fig4:image4}
  \end{minipage}\hfill
  \begin{minipage}{0.25\textwidth}
    \centering
    \includegraphics[width=\textwidth]{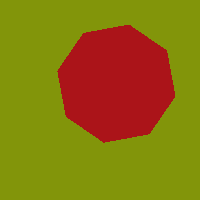}
    \caption{Octagon}{(Stop Sign)}
    \label{fig5:image5}
  \end{minipage}\hfill
  \begin{minipage}{0.25\textwidth}
    \centering
    \includegraphics[width=\textwidth]{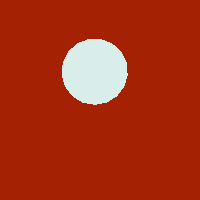}
    \caption{Circle}{(Speed Limit Sign)}
    \label{fig6:image6}
  \end{minipage}
\end{figure}

\subsubsection{Algorithm 5: Clean Shapes with Adversarial Road Sign Algorithm}
\hspace{1cm} The fifth algorithm extends the previous one by testing the model on adversarial road sign data, despite being trained on clean shapes data. This algorithm aims to assess the model's vulnerability to adversarial examples even when it has not been exposed to adversarial training during the training phase. The CNN architecture remains consistent with the previous algorithms. The training dataset consists of clean shapes data, while the testing dataset includes adversarial road sign images.

\subsubsection{Algorithm 6: Adversarial Shapes with Adversarial Road Sign Algorithm}
\hspace{1cm} The sixth algorithm explores the model's performance when trained on adversarial shapes data and tested on adversarial road sign data. This algorithm aims to assess the ability of the model to generalize and detect road signs effectively, even when faced with adversarial inputs that vary in terms of shape and color. The CNN architecture remains consistent with the previous algorithms, but the training dataset includes adversarial examples generated by altering the shapes of road signs while preserving their semantic meanings.

\hspace{1cm} It is important to note that under normal driving circumstances, Algorithm 1 (Control Algorithm) is expected to achieve the highest accuracy among the algorithms, as it is trained and tested on clean road sign data. Additionally, Algorithm 3 (Adversarial Road Signs with Adversarial Training Algorithm) is expected to outperform the other algorithms when facing adversarial attacks, thanks to the incorporation of adversarial training.

\subsection{Data Collection and Processing}
\hspace{1cm} The data collected for this study was acquired from Kaggle.com, a platform with a wide variety of public datasets relating to computer vision:
\begin{enumerate}[label=\Roman*\.)]
    \item Road Sign Detection - 877 images belonging to 4 classes \cite{roadsign}
    \item 2D Geometric Shapes Dataset - 520 images belonging to 4 classes \cite{shapes}
\end{enumerate}
\hspace{1cm} There were some challenges when it came to finding high-quality image data for United States road signs. Therefore, the road sign data that was used in this experiment originates from Russia and depicts common road signs in that area. The US road sign data is sparse in publicly available datasets. Although this does not alter the overall effectiveness of these models, it is important to note that collecting relevant and accurate data based on US road signs would have been more advantageous for training AV systems operating in the United States. The data was cleaned and carefully separated so that there were no duplicates or overlap between the training, validation, and testing sets, to avoid redundancy and possible overfitting because of faulty input data. The normalization process of these experiments go through the same basic premise:

\begin{enumerate}[label=\Roman*\.)]
    \item The image is loaded and resized to the target size of 224 x 224 pixels.
    \item The image is then converted into an array so that each element represents the pixel value of the corresponding location in the image.
    \item The pixel values are then normalized and divided by 255 in order to scale the values from the original range of [0, 255] to [0,1]. This is a typical step taken towards training machine learning models.
    \item An additional channel is created to incorporate a fourth class. This fourth class is needed to incorporate all 4 classes of this dataset: stop sign, speed limit sign, crosswalk sign, and traffic light.
    \item Random rotation is applied to rotate each channel by a random angle within the range of -90 to 90 degrees.
\end{enumerate}

\hspace{1cm} As this study mainly deals with the field of computer vision, and the goal of this study is to be able to accurately detect objects, the use of a convolutional neural network was the most beneficial deep learning model to use in this scenario. CNNs take the input of an image and produces its prediction of which class this particular image is a part of. They are designed for image analysis and due their convolutional layers, have the ability to processes features, patterns, edges, and textures in input images and can even be leveraged while implementing models for transfer learning; this can be particularly helpful in the case of this study.
\begin{quote}
\begin{enumerate}[label=\Roman*\.)]
    \item Control Algorithm (Clean Training + Clean Testing)
    \item Clean Training + Adversarial Testing
    \item Adversarial Training + Adversarial Testing
    \item Clean Shapes + Clean Road Sign Testing
    \item Clean Shapes + Adversarial Road Sign Testing
    \item Adversarial Shapes + Adversarial Road Sign Testing
\end{enumerate}
\end{quote}

\hspace{1cm} All models use the basis of a CNN with the same input layers, convolutional layers, pooling layers, and output layers. The datasets for the models with normal testing (1, 5) are split into 3 different sets with varying percentages from the original dataset: 70\% training, 20\% validation, and 10\% testing. The rest of the models (2, 3, 4, 6, 7) take a different approach, since they are are being tested and sometimes trained with adversarial examples, this split functions slightly differently. Whether the model is using adversarial training or adversarial testing, the split for training, validation, and testing is still 70\%, 20\%, 10\%, respectively, but instead of containing all clean image data, each of these categories are split in half, with one half being clean images and the other containing images with adversarial examples, or adversarial images. This assimilation of adversarial training as a defense technique against adversarial attacks should ensure a reasonable amount of accuracy when the system is faced with adversarial examples in real time, as opposed to the models that do not incorporate adversarial training. In the case of the models which are trained on shapes, the necessary and important features are extracted from the training images and used to detect their corresponding road signs. Each image has an associated annotation file which contains in order to teach the machine learning model which category of road sign each image is associated with; the ground truth labels for road signs include stop sign, speed limit sign, crosswalk sign, and traffic light.

\subsection{Experimental Setup \& Software}
This is the basic setup of the algorithms that will be compared and evaluated:
\begin{table}[H]
  \centering
  \includegraphics[width=0.9\textwidth]{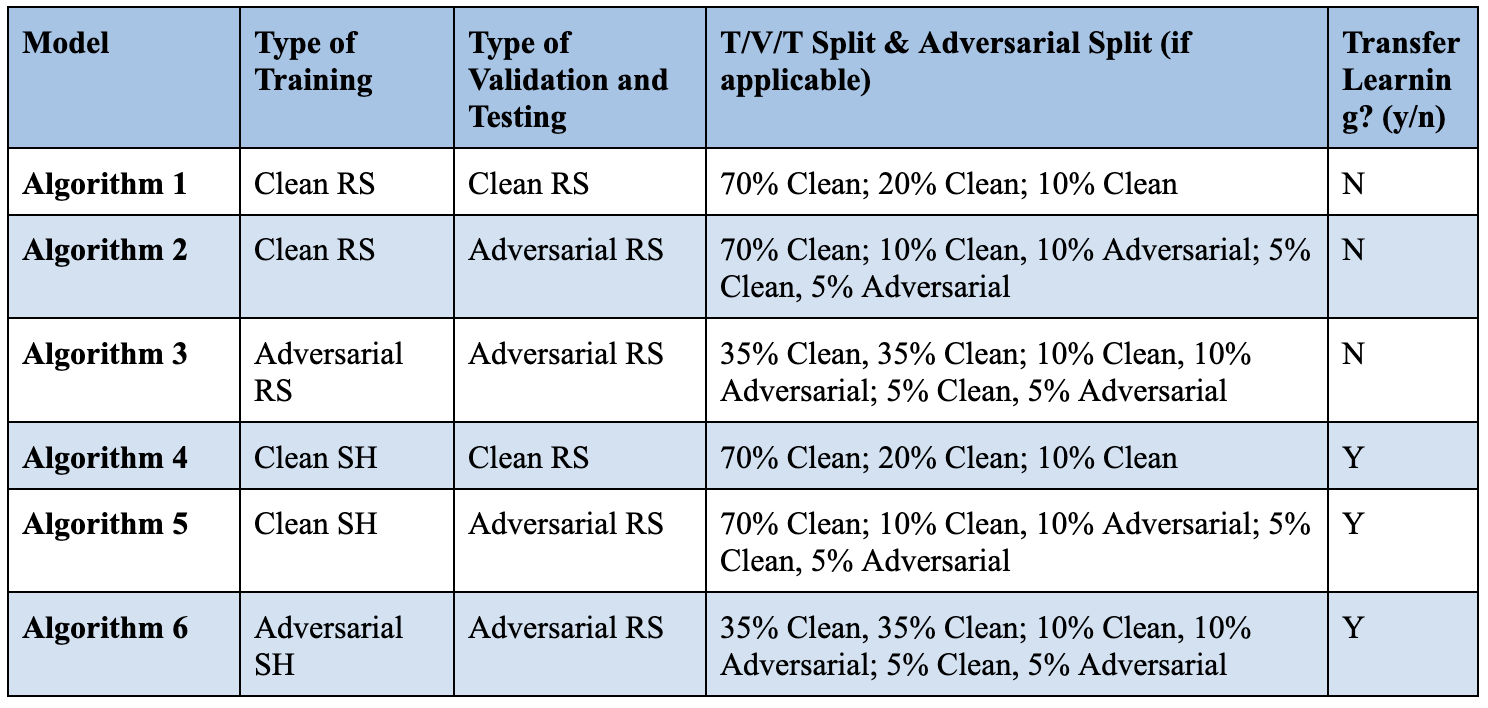}
  \caption{Detailed Descriptions of Algorithms}
  \label{tab1:image7}
\end{table}
\par
\hspace{1cm} In this project, several software tools and frameworks were utilized to facilitate the implementation and analysis of the experiments outlined above. All of the machine learning models were coded in Python using the Google Colab IDE, which allowed for the utilization of many deep learning frameworks and libraries that would prove useful when selecting, evaluating, and preprocessing models, designing layers of the convolutional neural network, preprocessing images, numerical computing, data manipulation, and plotting data points. The libraries used include: NumPy, pandas, sklearn (scikit-learn), Python Imaging Library (PIL). Multiple frameworks were strategically employed to aid in the development and training of the machine learning and deep learning models. PyTorch was utilized primarily for constructing and training neural networks. TensorFlow played a critical role in developing the machine learning models, leveraging its efficient computing capabilities for handling large datasets and complex architectures. To further streamline the process, Keras, operating on top of TensorFlow, provided an interface for building and training deep learning models. This seamless integration enhanced workflow efficiency, taking advantage of the strengths of the frameworks and libraries in synchrony.
\subsection{Statistical Analysis}
\hspace{1cm} The methods that were used to analyze and compare the models against each other were primarily focused on error analysis. The performance of the models during the training and validation processes are examined to determine their abilities to generalize input data, and therefore their abilities to synthesize and absorb the data as a whole. The curves displayed on these plots, also known as “learning curves”, represent the model’s performance. Further in-depth analysis will be performed in the Results section of this paper. These plots will be analyzed based on their robustness against adversarial attacks, computational efficiency, and ability to generalize. These will be calculated by how similarly they mimic the control model, Algorithm 1, and the amount of time they take to compute, respectively.

\section{Results}
\hspace{1cm} In this section, six figures will be presented, all corresponding to the results of their respective algorithms. Each algorithm was evaluated on 2 main datasets which were previously mentioned, Road Sign Detection \cite{roadsign} and 2D Geometric Shapes Dataset \cite{shapes}. The following plots and their training and validation losses will be evaluated to determine how effective they are compared to the original control algorithm which was trained and tested on clean road sign data.

\subsection{Experiment 1: Clean Road Sign Training with Clean Road Sign Testing (Control I) – Algorithm 1}
\begin{figure}[H]
  \begin{minipage}{0.6\textwidth}
    \centering
    \includegraphics[width=0.8\textwidth]{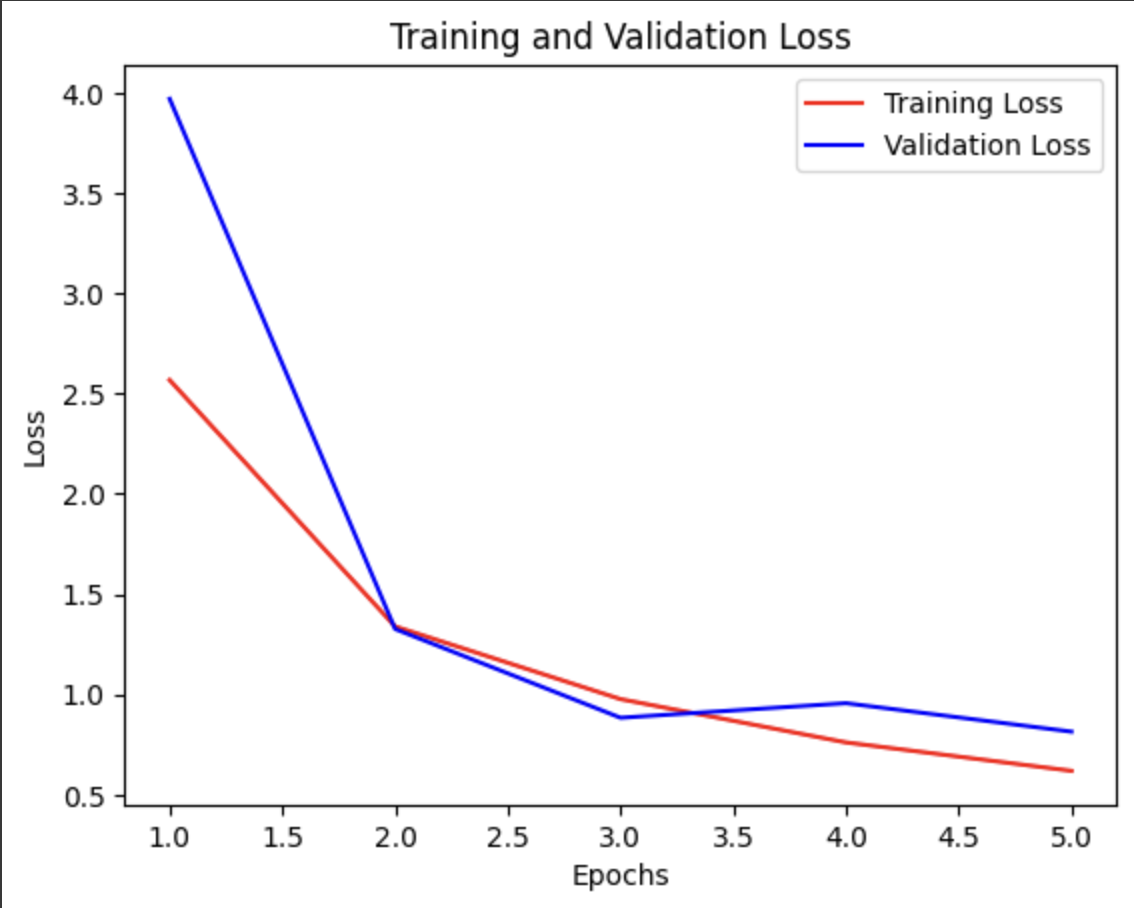}
    \caption{Training vs. Validation Loss for Algorithm 1}
    \label{fig7:image8}
  \end{minipage}%
  \begin{minipage}{0.4\textwidth}
    \begin{flushleft}
      This experiment's results during both the training phase and validation phase performed exactly as predicted. As evidenced by the decreasing trend in both training and validation loss over epochs. The quick drop to a low validation loss indicates poor generalization, which was also expected. Refer to Figure \ref{fig7:image8} for the plot of training vs. validation loss for Algorithm 1.
    \end{flushleft}
  \end{minipage}
\end{figure}

\subsection{Experiment 2: Clean Road Sign Training with Adversarial Road Sign Testing – Algorithm 2}
\begin{figure}[H]
  \begin{minipage}{0.6\textwidth}
    \centering
    \includegraphics[width=0.8\textwidth]{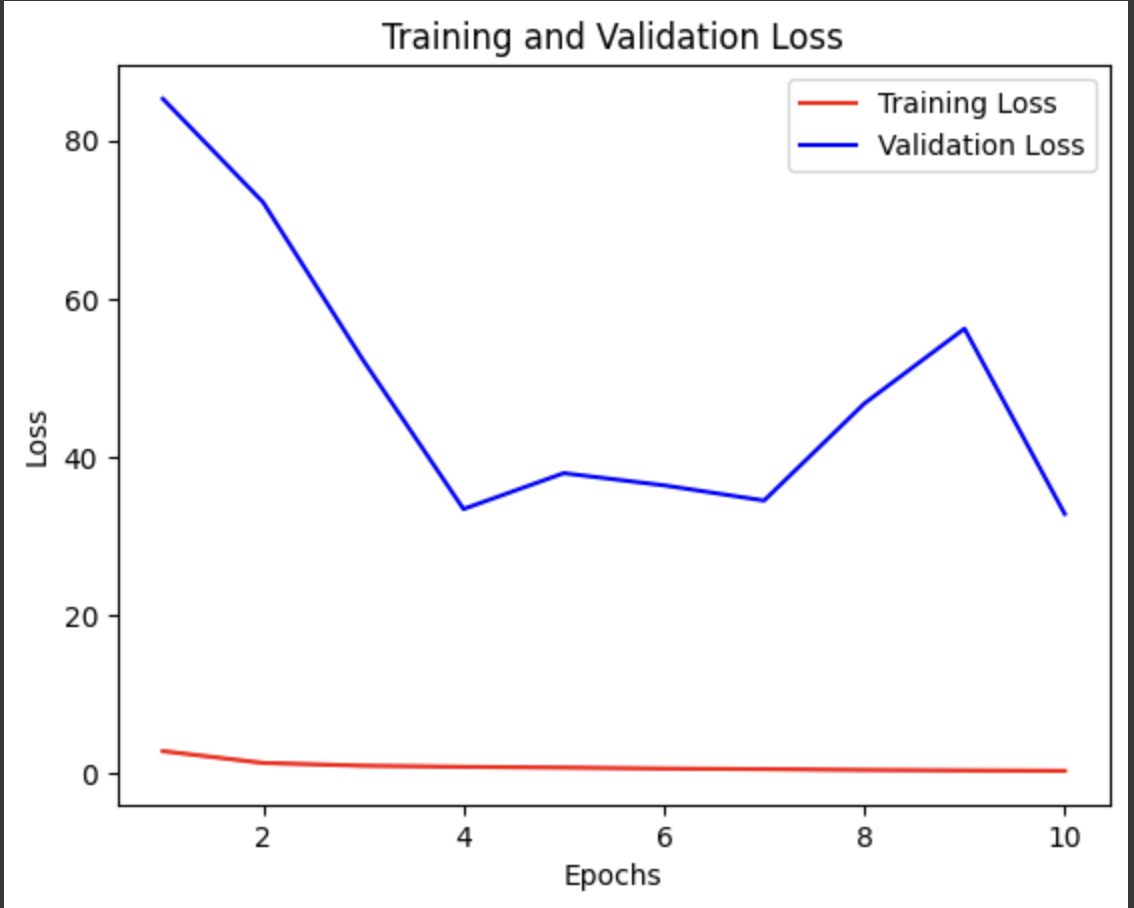}
    \caption{Training vs. Validation Loss for Algorithm 2}
    \label{fig8:image9}
  \end{minipage}%
  \begin{minipage}{0.4\textwidth}
    \begin{flushleft}
      Algorithm 2 exhibited different behavior during training compared to Algorithm 1. While the training loss held a steady low, the validation loss showed fluctuations and did not decrease at all. This is because the model was being tested on unseen adversarial road sign data, when it had only been tested on clean data. This behavior indicates underfitting, as the model struggled immensely to predict this unseen data. This demonstrates how an AV system could be expected to react when it has no exposure to adversarial examples during training: a system \textbf{without} adversarial training. See Figure \ref{fig8:image9} for the plot of training vs. validation loss for Algorithm 2.
    \end{flushleft}
  \end{minipage}
\end{figure}

\subsection{Experiment 3: Adversarial Road Sign Training with Adversarial Road Sign Testing – Algorithm 3}
\begin{figure}[H]
  \begin{minipage}{0.6\textwidth}
    \centering
  \includegraphics[width=0.8\textwidth]{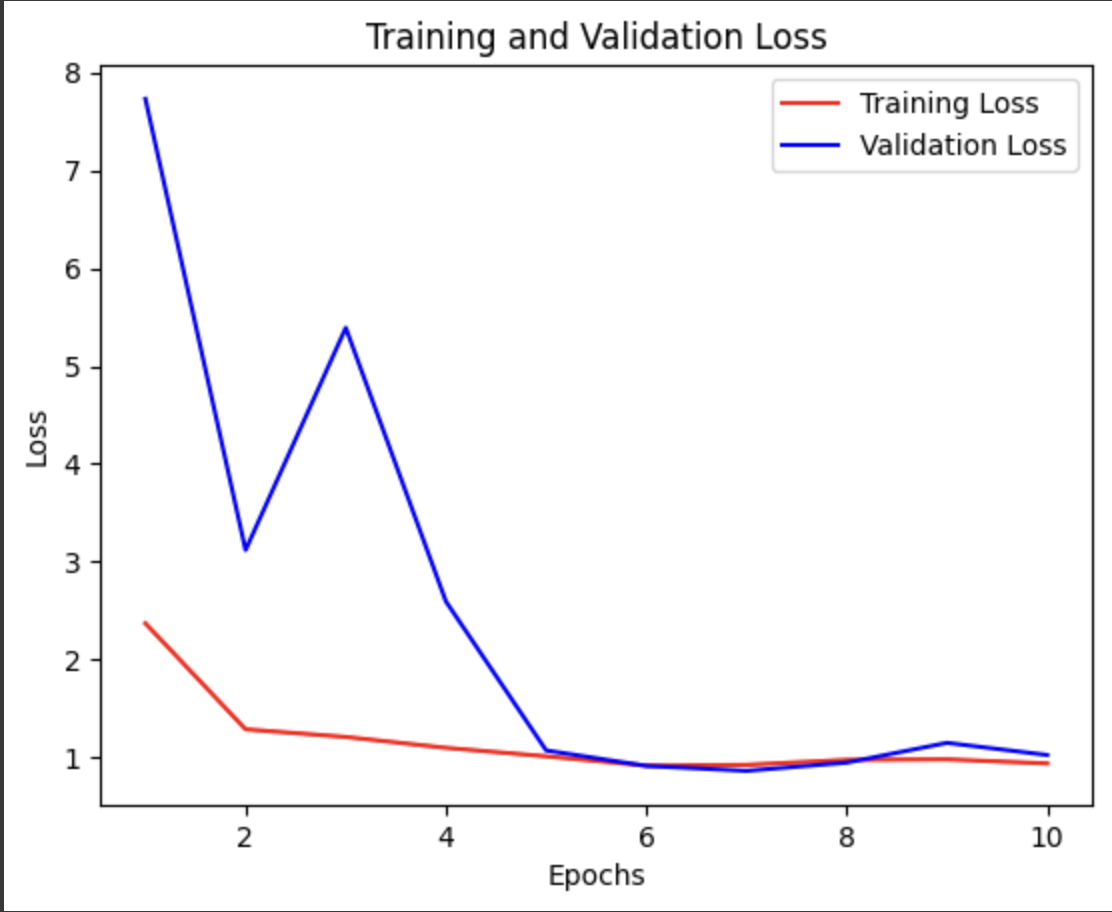}
  \caption{Training vs. Validation Loss for Algorithm 3}
  \label{fig9:image10}
  \end{minipage}%
  \begin{minipage}{0.4\textwidth}
    \begin{flushleft}
      Algorithm 3 demonstrated exceptional performance, with both the training and validation loss decreasing rapidly. The model achieved a remarkably low validation loss, indicating the model's ability to predict fairly accurately. This demonstrates how a model would function when it is exposed to adversarial examples during training: a system \textbf{with} adversarial training. The plot of training vs. validation loss for Algorithm 3 is below in Figure \ref{fig9:image10}.
    \end{flushleft}
  \end{minipage}
\end{figure}

\subsection{Experiment 4: Clean Shapes Training with Clean Road Sign Testing (Control II) – Algorithm 4}
\begin{figure}[H]
  \begin{minipage}{0.6\textwidth}
    \centering
  \includegraphics[width=0.8\textwidth]{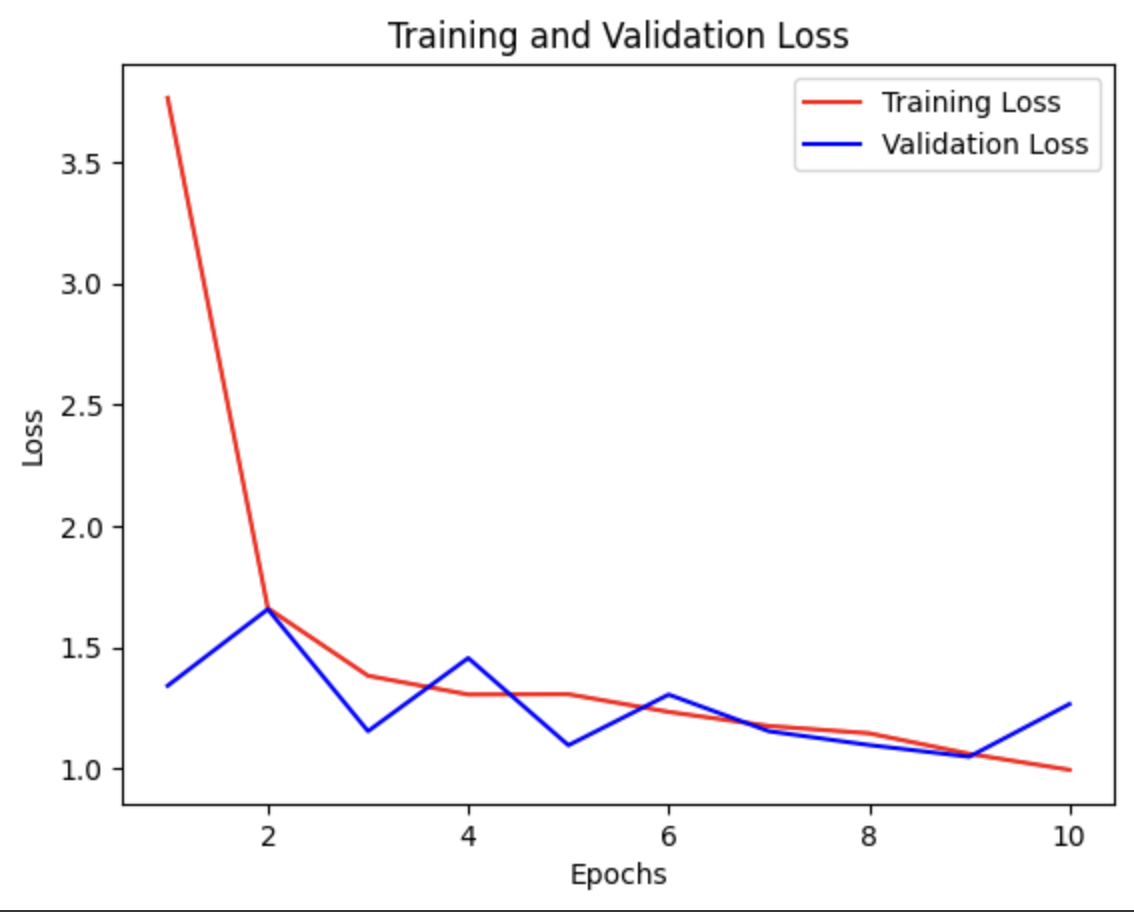}
  \caption{Training vs. Validation Loss for Algorithm 4}
  \label{fig10:image11}
  \end{minipage}%
  \begin{minipage}{0.4\textwidth}
    \begin{flushleft}
      Algorithm 4 displayed an interesting behavior during validation. The training loss decreased slowly and plateaued later on, while the validation loss started low and fluctuated throughout the epochs. This behavior suggests that the model generalizes well, as it has only been exposed to clean shape data and is being tested on clean road signs. It could benefit from further regularization to achieve better and smoother results. Refer to Figure \ref{fig10:image11} for the plot of training vs. validation loss for Algorithm 4.
    \end{flushleft}
  \end{minipage}
\end{figure}

\subsection{Experiment 5: Clean Shapes Training with Adversarial Road Sign Testing – Algorithm 5}
\begin{figure}[H]
  \begin{minipage}{0.6\textwidth}
    \centering
  \includegraphics[width=0.8\textwidth]{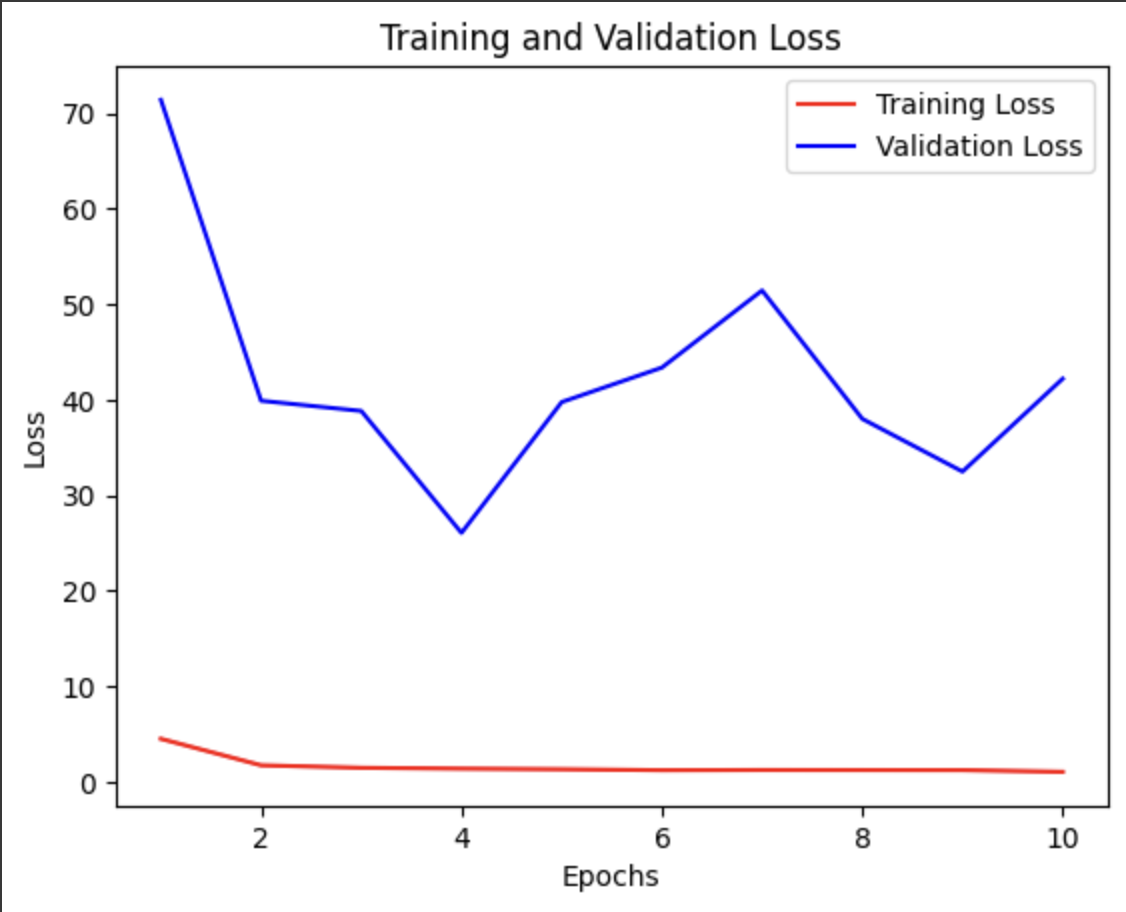}
  \caption{Training vs. Validation Loss for Algorithm 5}
  \label{fig11:image12}
  \end{minipage}%
  \begin{minipage}{0.4\textwidth}
    \begin{flushleft}
     Algorithm 5 showed performed exactly as it was expected to. Since it was trained on clean shape data and only exposed to this initially, when it was tested on adversarial road sign data, it did not know how to classify the objects in the images. This figure depicts incredible underfitting, as the figure for Experiment 2 did as well. Further investigation as to why this similarity exists will be discussed later on. See Figure \ref{fig11:image12} for the plot of training vs. validation loss for Algorithm 5.
    \end{flushleft}
  \end{minipage}
\end{figure}

\subsection{Experiment 6: Adversarial Shapes Training with Adversarial Road Sign Testing – Algorithm 6}
\begin{figure}[H]
  \begin{minipage}{0.6\textwidth}
    \centering
  \includegraphics[width=0.8\textwidth]{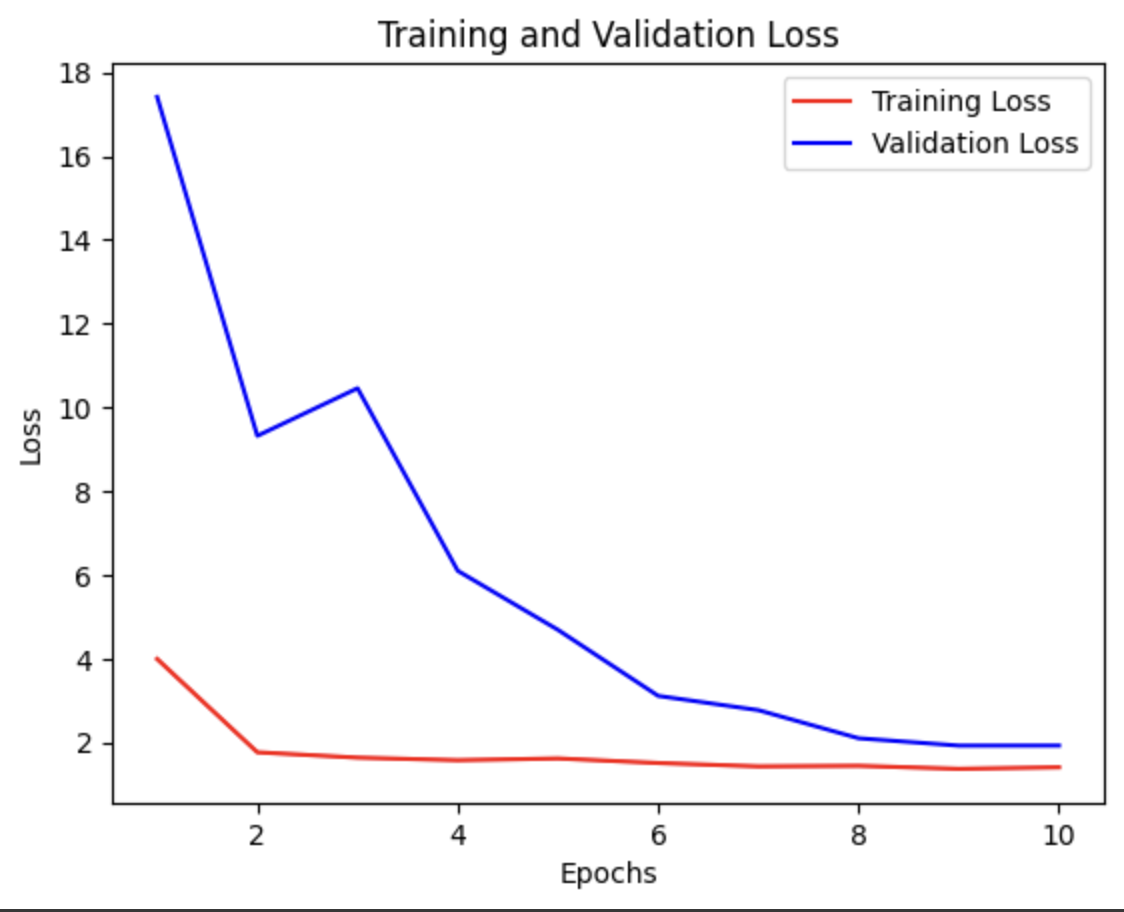}
  \caption{Training vs. Validation Loss for Algorithm 6}
  \label{fig12:image13}
  \end{minipage}%
  \begin{minipage}{0.4\textwidth}
    \begin{flushleft}
     Algorithm 6 exhibited slow convergence during training and validation, with both the training and validation loss decreasing gradually. The model achieved a reasonably low validation loss, which is promising, as it does not indicate any signs of overfitting nor underfitting. The plot of training vs. validation loss for Algorithm 6 is shown in Figure \ref{fig12:image13}. These results will be further discussed and analyzed in the next section.
    \end{flushleft}
  \end{minipage}
\end{figure}

\section{Discussion}
\hspace{1cm}Since the results were discussed in the previous section, there will be an overall analysis and comparison completed in this section, as well as the comparison of some algorithms and their performance in more detail. The overall comparison of the computational efficiency, average accuracy, and generalizability of each algorithm are as follows:

\begin{figure} [H]
  \centering
  \begin{minipage}{0.5\textwidth}
    \centering
    \includegraphics[width=\textwidth]{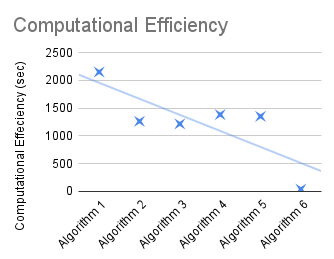}
    \caption{Computational Efficiency}
    \label{fig13:image14}
  \end{minipage}\hfill
  \begin{minipage}{0.5\textwidth}
    \centering
    \includegraphics[width=\textwidth]{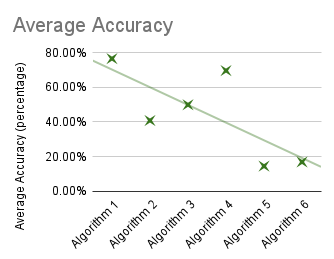}
    \caption{Average Accuracy}
    \label{fig14:image15}
  \end{minipage}\hfill
  \begin{minipage}{0.5\textwidth}
    \centering
    \includegraphics[width=\textwidth]{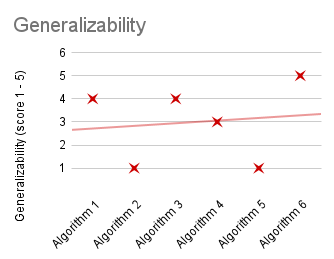}
    \caption{Generalizability}
    \label{fig15:image16}
  \end{minipage}
\end{figure} 

\par \hspace{1cm} In the evaluation of the various algorithms, I encountered a complex array of trade-offs between computational efficiency, accuracy, and generalization capabilities. These algorithms shared a common architecture to maintain consistency across variables, allowing me to focus on the interplay of these fundamental performance metrics. To synthesize and derive meaningful insights from this diverse set of data, I employed Multiple Criteria Decision Analysis (MCDA), which will be discussed later in this section.

\hspace{1cm} In order to discern the strengths and weaknesses of each algorithm, MCDA provided a structured approach to weigh and evaluate various performance metrics – computational efficiency \ref{fig13:image14}, average accuracy \ref{fig14:image15}, and generalizability \ref{fig15:image16}. MCDA is a statistical test that enables the comprehensive assessment and ranking of alternatives based on multiple criteria. By assigning equal weights to these criteria, I applied MCDA to calculate an overall score for each algorithm, which allowed for the objective ranking of them. If, instead, there was a need to give more weight to one category over the others, MCDA would enable the calculation of that as well. The pseudocode for the Python implementation of how MCDA was used to analyze and rank the algorithms as well as an analysis of its results is presented below.

\begin{algorithm}[H]
\caption{Ranking Algorithms Using MCDA}
\begin{algorithmic}
\State
\State \textbf{Input:} Data for each algorithm, Criteria weights
\State \textbf{Output:} Ranked algorithms
\State
\State \# define the data for each algorithm:
\State \hspace{0.5cm} - Computational Efficiency
\State \hspace{0.5cm} - Average Accuracy
\State \hspace{0.5cm} - Generalizability
\State 
\State \# define the weights for the criteria:
\State \hspace{0.5cm} - \textit{Weight\_Efficiency} = 1/3
\State \hspace{0.5cm} - \textit{Weight\_Accuracy} = 1/3
\State \hspace{0.5cm} - \textit{Weight\_Generalizability} = 1/3
\State 
\State \# create a dictionary to store the overall scores:
\State \hspace{0.5cm} \textit{scores} = \{\}
\State 
\For{each algorithm in the data}
\State \# calculate normalized values for each criterion:
\State \hspace{0.5cm} - \textit{Normalized\_Efficiency} = (Efficiency - Minimum Efficiency) / (Maximum Efficiency - Minimum Efficiency)
\State \hspace{0.5cm} - \textit{Normalized\_Accuracy} = Accuracy / Maximum Accuracy
\State \hspace{0.5cm} - \textit{Normalized\_Generalizability} = (Generalizability - Minimum Generalizability) / (Maximum Generalizability - Minimum Generalizability)
\State 
\State \# calculate the overall score for the algorithm:
\State \hspace{0.5cm} - \textit{Overall\_Score} = (\textit{Weight\_Efficiency} * \textit{Normalized\_Efficiency}) + (\textit{Weight\_Accuracy} * \textit{Normalized\_Accuracy}) + (\textit{Weight\_Generalizability} * \textit{Normalized\_Generalizability})
\State 
\State \# store the Overall Score in the dictionary with the algorithm's name as the parameter
\State \hspace{0.5cm} \textit{scores[algorithm]} = \textit{Overall\_Score}
\State \textbf{end for}
\EndFor
\State 
\State \textbf{Output:} Ranked algorithms based on MCDA
\State
\State \# sort the dictionary of overall scores in descending order to rank the algorithms
\For{rank, algorithm in enumerate(ranked\_algorithms)}
\State \# print the formatted ranked algorithms
\State \hspace{0.5cm} - \textit{print(f"{rank}. {algorithm}: Overall\_Score = {scores[algorithm]:.4f}")} \normalfont
\State \textbf{end for}
\EndFor
\State
\end{algorithmic}
\end{algorithm}

\par The outcomes presented below are the outputs of executing this Python code on the dataset across each algorithm:
\begin{enumerate}
\item \textbf{Algorithm 1}: Overall Score $\approx$ 0.9167
\item \textbf{Algorithm 4}: Overall Score $\approx$ 0.6821
\item \textbf{Algorithm 3}: Overall Score $\approx$ 0.6532
\item \textbf{Algorithm 6}: Overall Score $\approx$ 0.4070
\item \textbf{Algorithm 2}: Overall Score $\approx$ 0.3708
\item \textbf{Algorithm 5}: Overall Score $\approx$ 0.2703
\end{enumerate}

\par \hspace{1cm} Among the evaluated algorithms, \textbf{Algorithm 1}, trained and tested on clean road sign data, exhibited the least favorable computational efficiency, while having the highest accuracy. Despite this proficiency in accuracy, its generalizability stood at 80\%, implying that it may face challenges when presented with unseen data outside the training distribution, but still sound enough to generalize adequately. Despite this, it was ranked as the best algorithm, with an Overall Score of 92\%, which again, is expected from this control algorithm. \textbf{Algorithm 2}, on the other hand, demonstrated average computational efficiency, accuracy, and poor generalizability. The fact that this model was tested on data that it had not yet been exposed to suggests a limited capacity to generalize to diverse scenarios. Overall, this is a poor model and should be used as a benchmark for how models should not perform, with being ranked the second worst algorithm and having the MCDA Overall Score of 37\%. \textbf{Algorithm 3} displayed moderate computational efficiency, a slightly above-average accuracy, being third in ranking, and having the Overall Score of 65\%. Similar to Algorithm 1, its generalizability also reached 80\%. Although its accuracy did not surpass that of Algorithm 1, the results indicate a consistent ability to perform well on diverse datasets, since this was the model that was trained and tested on adversarial data.

\par \hspace{1cm}\textbf{Algorithm 4} displayed average computational efficiency, performing slightly above average in accuracy while exhibiting typical generalizability, being ranked the second best algorithm. Its above-average performance indicates that it is a viable option for use in certain situations; further optimization should be considered for higher accuracy, but its Overall Score still indicates 68\%. \textbf{Algorithm 5} demonstrated average computational efficiency, but its accuracy performance was relatively low, and its generalizability was notably inadequate, being ranked as the worst algorithm with an Overall Score of 27\%. The limited generalization capability raises concerns about real-world applications and highlights the necessity for improved training strategies. This was tested on adversarial data, but trained on clean data, which explains its inability to generalize effectively. Finally, \textbf{Algorithm 6}, exhibited the best computational efficiency, the second-poorest accuracy, and the best generalizability, which brought its ranking to number 4 with an Overall Score of 41\%. Leveraging transfer learning and human perception mechanisms, it proved to be an effective solution for this environment, showcasing robustness and adaptability to diverse real-world scenarios. The accuracy can be further tested and improved with the fine-tuning of the model.

\par \hspace{1cm} The comprehensive evaluation of these algorithms highlights the trade-offs between computational efficiency, accuracy, and generalization capabilities. Since all of the algorithms utilized the same overall architecture in order to keep these variables constant, further studies can fine-tune these models to test how well they can perform with different convolutional layers and regularization. This will be necessary to take these experiments and elevate them beyond just baseline testing. If these algorithms were re-tested after optimizing the convolutional layers and regularization strategies to best suit each one, it is likely their average accuracies could dramatically increase.

\par \hspace{1cm} The utilization of MCDA allowed me to conduct a systematic and impartial evaluation of these algorithms, considering all three essential criteria simultaneously. This approach aids in decision-making by providing a clear ranking of the algorithms based on their overall performance, considering the trade-offs between computational efficiency, accuracy, and generalizability. Transfer learning played a crucial role in the success of this experiment. Because of transfer learning, the models in the last three experiments (\textbf{Algorithms 4, 5, and 6}) were able to produce strong results, despite being trained on shape data and being tested on road sign data, without any prior exposure. The model was able to learn from the original shape training data and applied its knowledge to the different but related road sign testing data. The combination of MCDA and transfer learning facilitated the insights into the performance of these algorithms, paving the way for future optimization and fine-tuning to further enhance their capabilities.

\par \hspace{1cm} Transfer learning was especially useful for this study because of the limited data available for road signs, feature extraction, regularization, and avoiding data annotation. Using the shape dataset allowed for the extension of data to be used for these models. They were able to leverage the knowledge from the adversarial shape data to improve the model's performance on the adversarial road sign data, even with the limited data available. During the training period on the shape data, the model learned useful features that are pivotal for the road sign data as well. These lower-level features capture general patterns and visual representations that can be shared across related tasks, i.e. their shapes and colors. Transfer learning also acted as a more enhanced for of regularization, as it prevented overfitting on the road sign data, therefore increased the model's ability to generalize the input data and avoid memorizing irrelevant patterns in the road sign data. Lastly, since the road sign dataset was already annotated, but the shape dataset was not, transfer learning eliminated the need to manually annotate the shape dataset, saving time.

\section{Future Works}
\hspace{1cm} While this study has made an effort to contribute to the ongoing research in physical adversarial attacks on autonomous driving systems by proposing methods to strengthen object classifiers, there still exist many potential avenues for future research and investigation. The following are a set of key ideas that could enhance the scope and impact of this study:

\begin{quote}
\subsection{Human-in-the-Loop Validation}
\hspace{1cm} This research focuses on strengthening algorithms, models, and human-vehicle interfaces, a critical aspect that warrants future exploration is human-in-the-loop validation. In this study, I implemented an aspect of human-in-the-loop validation where the user would be able to assist in classifying road signs that the system was unsure about. The ethical implications of incorporating human input into the training data raise concerns about potential biases or inaccuracies in the model. Future works should delve into the security risks associated with human-computer interactions and propose methods to validate and ensure the integrity of the added human input. The flowchart for the chain of events depicted below:
\begin{figure}[H]
  \centering
  \includegraphics[width=1\textwidth]{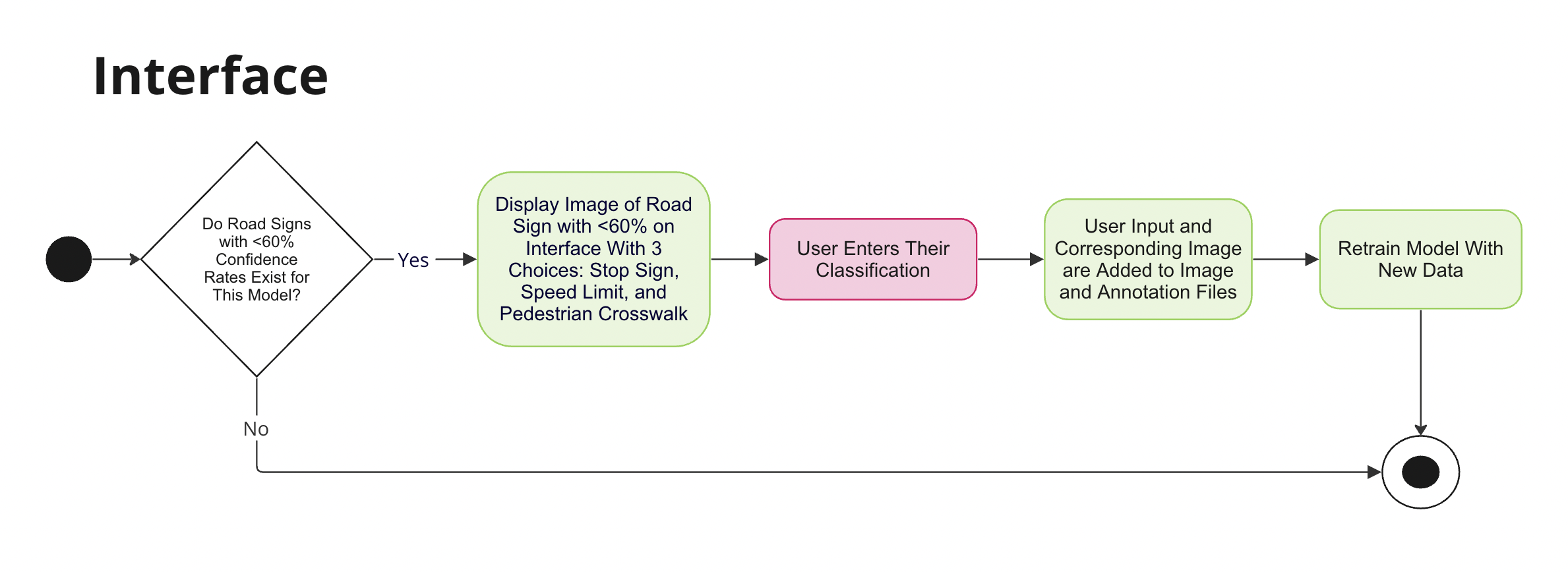}
  \caption{Interface Flowchart}
  \label{fig16:image17}
\end{figure}

\subsection{Security and Risk Analysis}
\hspace{1cm} To provide a more holistic investigation of the security of autonomous driving systems, future research should expand its scope to include an analysis of security risks beyond physical adversarial attacks. Investigating potential vulnerabilities in digital adversarial attacks and system-level attacks would contribute to a more comprehensive understanding of autonomous vehicle security.

\subsection{Real-World Testing}
\hspace{1cm} Moving beyond simulation-based evaluations, future works should explore opportunities for real-world testing and deployment of the proposed algorithms and models. Of course, this would mean gathering more effective, relevant, and significant data to train these systems with. Conducting experiments on actual autonomous vehicles in controlled environments can provide valuable insights into the algorithms' real-world applicability.

\subsection{Integration of Explainable AI} 
\hspace{1cm} As deep learning algorithms become technically complex, understanding the decision-making process of AVs becomes crucial, especially in safety-critical applications, such as autonomous driving. Future research should consider integrating explainable AI techniques. 
As mentioned in \textit{Human drivers’ situation awareness of autonomous driving
under physical-world attacks}, "Explainable AI (XAI) is a research
field that aims to make AI systems’ decisions more transparent
and understandable to humans" \cite{aiping} \cite{mueller}. This integration would make AV decisions more transparent to the end user and to provide interpretable insights into the model's decisions.
\end{quote}
Pursuing these future directions could significantly advance the field and enhance autonomous vehicle safety and security in real-world driving scenarios.

\section{Acknowledgments}
As the author of this paper, I would like to express my gratitude by acknowledging the following individuals and organizations for their support on this research:

\begin{itemize}
  \item Dr. Aiping Xiong, Principal Investigator, for her mentorship.
  \item Katherine Zhang, Graduate Student Advisor, for her valuable insights.
  \item The Pennsylvania State University for providing research facilities and resources.
  \item The National Science Foundation for funding this research.
\end{itemize}

This research would not have been possible without the support and contributions of the above individuals and organizations.

\bibliographystyle{unsrt}  
\bibliography{references}

\end{document}